\documentclass{article}

\usepackage[a4paper, left=1.5in, right=1.5in, top=1in, bottom=1in]{geometry}

\usepackage{flushend}

\usepackage[english]{babel}
\usepackage{verbatim}
\usepackage[autostyle]{csquotes}
\usepackage[ruled]{algorithm2e}

\usepackage{multirow}
\usepackage{amsmath}
\usepackage{array}
\usepackage{booktabs}
\usepackage{graphicx}
\usepackage{siunitx}
\usepackage{adjustbox}
\usepackage{subcaption}
\usepackage{pgfplots}
\pgfplotsset{compat=1.16}

\usepackage{colortbl}
\usepackage{xcolor}

\usepackage{algorithm}
\usepackage{algpseudocode}
\usepackage{amsmath}
\usepackage{amsfonts}
\usepackage{booktabs}
\usepackage{array}
\usepackage{graphicx}
\usepackage{caption} 
\usepackage{bm}
\usepackage{hyperref}
\title {A Keyword-Based Technique to Evaluate Broad Question Answer Script}

\author{Tamim Al Mahmud$^{*}$, Md Gulzar Hussain, Sumaiya Kabir,\\ Hasnain Ahmad, Mahmudus Sobhan \\
        \small Department of Computer Science and Engineering\\ Green University of Bangladesh\\ Dhaka, Bangladesh \\
\\\\
        \small $^{*}$Corresponding author: Tamim Al Mahmud; \tt{tamim@cse.green.edu.bd} \\
}
\date{}

\begin{document}
\maketitle
\begin{abstract} 
  Evaluation is the method of assessing and determining the educational system through various techniques such as verbal or viva-voice test, subjective or objective written test. This paper presents an efficient solution to evaluate the subjective answer script electronically. In this paper, we proposed and implemented an integrated system that examines and evaluates the written answer script. This article focuses on finding the keywords from the answer script and then compares them with the keywords that have been parsed from both open and closed domain. The system also checks the grammatical and spelling errors in the answer script. Our proposed system tested with answer scripts of 100 students and gives precision score 0.91.

\end{abstract}

\vspace{1em}

\section{Introduction}\label{introduction}
Digitally evaluation is the process of collecting, analyzing, and interpreting information to determine the extent of the students are achieving the instructional objective. Every educational institute evaluates the answer scripts of the students to assess a student’s performance. There are several types of test to evaluate student's performance such as verbal or viva-voice, subjective or objective. Currently, the evaluation of multiple choice question (MCQ) tests are done by using the optical mark reader (OMR) machine. It is very profitable to lessen the use of resources, but this approach only applies to evaluate the MCQ answer scripts only. The descriptive test is helpful to find out the depth knowledge of a student on a particular course. In term of the broad question, a student has to describe the answers of a particular question from one to many sentences. To evaluate a large number of those answer scripts manually, the evaluators face a unanimous amount of work pressure. If it is possible to evaluate subjective or descriptive answer scripts electronically then it will be a great achievement for the education system. It will save time, reduce resource utilization, prejudice mistakes of the evaluators. In term of the recruitment process, the organizations usually evaluate the candidate's ability through the written test. Each year the number of applicants in each job sector is increasing dramatically. As a result, the number of the subjective answer script is also increasing. In this paper, we have demonstrated and implemented an integrated software system to evaluate an answer script electronically.

The primary objectives of our work are-
\begin{itemize}
\item building an efficient system that will evaluate broad question answer script electronically.
\item providing the facility of open and close domain question answering. 
\item shaping an error free system for evaluation. 
\end{itemize}

The remaining of the article is structured as follows. Related works are discussed in \textbf{Section~\ref{sec:related}}. \textbf{Section~\ref{sec:method}} discusses methodology and a sample data is illustrated and \textbf{Section~\ref{sec:res}} shows the results. Finally \textbf{Section~\ref{sec:con}} refers the future work and conclusion.

\section{Background Overview}\label{sec:related}
In this section, we have discussed briefly some of existing works that are related with our work. 

Various techniques to evaluate free text or subjective answers automatically like Latent Semantic Analysis (LSA), Intelligent Essay Assessor, Syntactically Enhanced LSA (SELSA) are reviewed by the authors of paper \cite{rinchen2014comparative} and found that these approaches are keyword oriented and didn't think about the adjacent terms. Authors of \cite{hussain2019assessment} proposed a model to evaluate Bangla descriptive answer script where they found minimum relative error of 1.8\%. Paper \cite{chen2017reading} suggests addressing open-domain answering questions using Wikipedia as the distinctive source of information. In this paper \cite{chen2016using}, the authors used a method that uses a coarse-grained answer type. In this paper, they presented a novel automatic question answering system being the first research to deal with various types of user questions in the consumer electronics domain \cite{yoon2016automatic}. In their work the authors of \cite{brzeski2014relation} proposed an alternative keywords search method for Wikipedia, designed as a model solution for answering factual questions. This paper presents a new system that can evaluate student’s performance at higher level considering the assessment of broad type questions by parsing of text and find the semantic meaning of student's answer and finally compare it with instructor's answer and assigns the final scores \cite{patil2014intelligent}. The author of \cite{lcock2012wikitalk} proposed an open-domain information access system that uses Wikipedia articles as its source of information to discuss subjects. In \cite{islam2010automated} the authors developed an aging system using generalized latent semantic analysis (GLSA) that can achieve the higher level of performance than a human grader. In paper \cite{moneautomatic}, the authors presented a system that attempts to solve both close and open domain problem where open domain problems can be solved by using a search engine and answers of the closed domain problem have to be stored in a database. In \cite{sasaki2005overview} the authors describes about the NTCIR who deal with multi-lingual question answer in Chinese and Japanese. Wikipedia-based question answer is also a hot topic. In \cite{buscaldi2006mining}, they treated Wikipedia as a knowledge base.

In \cite{radev2002evaluating} the author discussed NSIR (pronounced ''Answer''), a web-based question answering mechanism that was in the process of development at University of Michigan. Once NSIR receives the search engine's hit list, it processes the high-ranked records and extracts a number of prospective responses. Authors of \cite{tulaskar} proposed a system that will take the answer as input then compare it with the stored standard answer and evaluate each reply by matching keywords or main ideas with conventional reply synonyms.  In \cite{pulman2005automatic}, this paper their the objective was to explore the methods of computational linguistics in automatically marking free text answers.

In \cite{mittal2005fully} this work the authors presented a fully automated question and answering system for intelligent search in e-learning documents. The method utilizes the processing methods of natural language to define the semantic and syntactic structure of the issue. Lastly \cite{xu2008research}, this paper presents the application framework and theory, that are built on question semantic depiction and ontology. It also presents the important techniques: question parsing, knowledge base construction, answer extraction.

\section{Methodology}\label{sec:method}
In this section, we discuss the working methodology of our proposed system.
\\
Our system takes two steps to evaluate the answer script of the examinee which are given in Figure 1.

\begin{figure}[htb!]
	\centering
	\includegraphics[width=8.5cm,height=4cm]{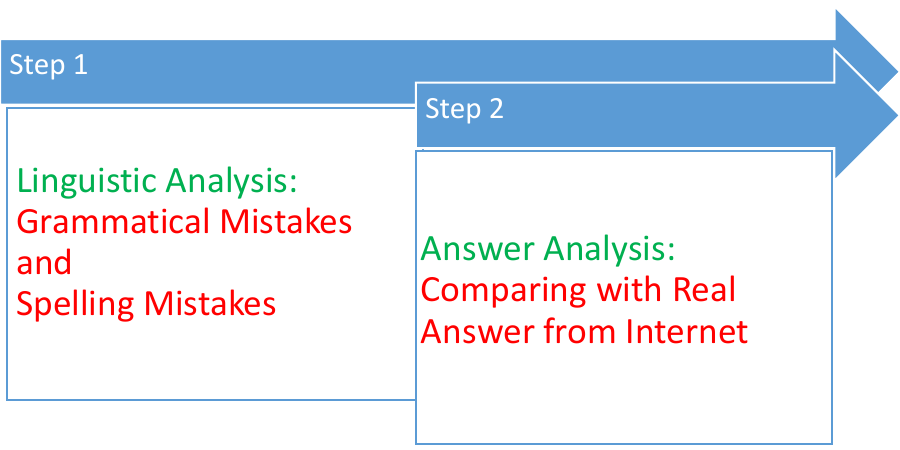}
	\caption{Steps of Evaluating Answer Script}
	\label{fig:fig1}
\end{figure}

The first step is Linguistic Analysis, to check Grammatical Mistakes and Spelling Mistakes in the answer script. For the second step Answer Analysis, we proposed two algorithms to find the frequency of words in the answer script of the examinee and the result from the open domain and closed domain. These steps are discussed below:

\subsection{Linguistic Analysis}
We can use tools like jlangualetool, Perfect Tense API, Grammar Bot API for spell checking and grammar checking. The reason is, these are easy and free to use language tool as a spell and grammar checker in our projects while we can also use existing projects like hunspelljna or hunspellbird ( at maven central). These are powerful grammatical correction tools which understands the meaning and context accurately of text.

Using these tools, to give a linguistic score we proposed Algorithm 1-

\begin{algorithm}[H]\label{algo2}
\SetAlgoLined
String answer $=$ Total answer of the student\;\\
Initial LAScore $=$ 0\;\\
Initial SMistake $=$ Number of spelling mistakes\;\\
Initial GMistake $=$ Number of grammatical mistakes\;\\
Initial TWord $=$ Number of Words in answer\;\\
Initial TSentence $=$ Number of Sentence in answer\;\\
$LAScore = \frac{SMiskate}{TWord}*100 + \frac{GMiskate}{TSentence}*100$

\caption{Algorithm to Generate Linguistic Score}
\end{algorithm}

\subsection{Answer Analysis}
The system architecture of the answer analysis step is given in Figure 2. 

\begin{figure}[htb!]
	\centering
	\includegraphics[width=8.5cm,height=12.5cm]{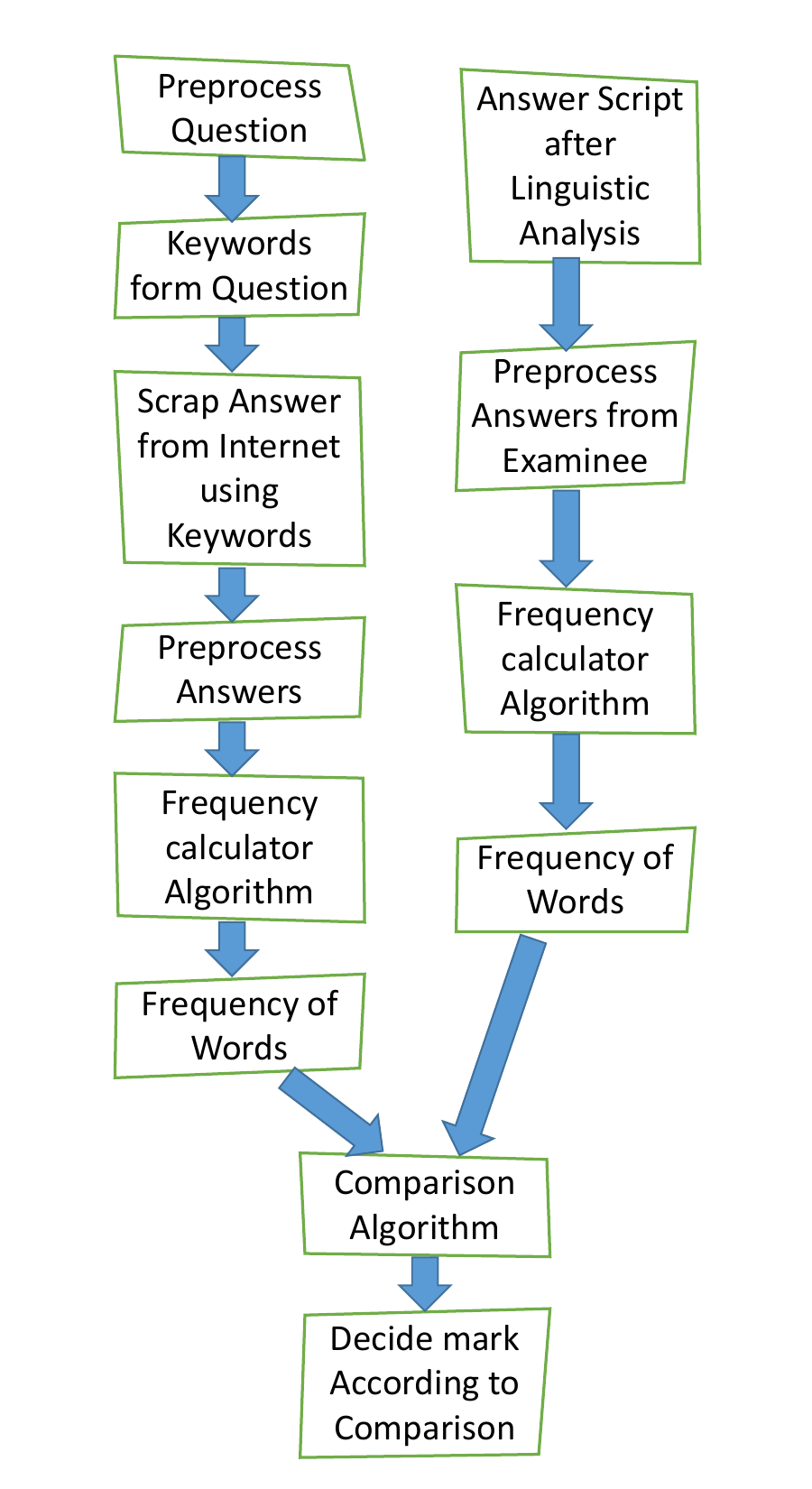}
	\caption{System Architecture of the Answer Analysis}
	\label{fig:fig2}
\end{figure}

Required process of this step is elaborated below:

\subsubsection{Scrapping Answers from Domains}
Our proposed system will focus on open domain system after extracting the keyword from the question then the system will parse data from the open domain like Wikipedia. The Mediawiki action API is a web service that provides convenient access to wiki features.
Entry point: \enquote{https://en.wikipedia.org/w/api.php}, or any other wiki. Parameters are passed in the query string. Null passing will give us the help page with the auto-generated documentation. We need to select an output format. The MediaWiki offers data output in the form of JSON, jsonfm, php (serialized format), phpfm, wddx, wddxfm, XML, xmlfm, yaml, yamlfm, and rawfm. The formats suffixed "fm" are pretty-print in html. This paper considers the problem of answering questions in an open-domain setting using Wikipedia. Wikipedia contains updated knowledge that humans are interested in. The main objective of this system is to retrieve the correct answer to the queries posed by the users. The returned answer is in the form of relevant documents. Data source maybe world wide web or local domain system, it will take a question in a set of keywords, for example, what is the open domain system? It will accept all the data about this keyword from Wikipedia then it reduces the unnecessary part of the parsed data using preprocessing.

\subsubsection{Preprocessing}
In this process whenever a text is rendered, specific words are extracted. Stop words like a, and, the, is, etc are removed during this process. Normalization process that is removing unnecessary characters and special characters like '\#', ',' etc is also happened in this process. Capitalization is removed like 'Hello' is converted to 'hello'. But some examples like 'US' isn't converted to 'us' as it changes the meaning. Tokenization is also used to split paragraphs in to sentences, and sentences in to single words. These preprocessing are done using various regular expressions(RE).  

\subsubsection{Word Frequency Calculator}
We proposed an algorithm to calculate the frequency of every words after preprocessing the answer. This process happens for both of the answer script from examinee and scrapped result from domains. For this Algorithm 2 is proposed-

\begin{algorithm}[H]\label{algo2}
\SetAlgoLined
String answer $=$ Total answer after preprocessing\;\\
Initial Array frequency $=$ null\;\\
\For{Every words in answer}
{
  \eIf{Word is not in the Array}{
   Add the word as index in the Array\;\\
   frequency[word] $=$ 1\;
   }{
   frequency[word] $=$ frequency[word] $+$ 1\;
  }
}
\caption{Word Frequency Calculator Algorithm}
\end{algorithm}

\subsubsection{Comparison Between Two Results}
After running the word frequency calculator algorithm on both of the result of parsed data from open and closed domain we will get two frequency set. To evaluate the answer of the student we have to compare between the two frequency sets and have to set score on the comparison result. For doing this we proposed Algorithm 3-

\begin{algorithm}[H]\label{algo8}
\SetAlgoLined
Initial AAScore $=$ 0\;\\
$SWFrequency = Frequency Result Of Student Answer$\;\\
$RWFrequency = Frequency Result Of Parsed Answer$\;\\
Initial lengthSWF $=$ Length of SWFrequency\;\\
Initial lengthRWF $= \sum All Values Of RWFrequency$\;\\
Initial WeithtRWF $=$ Empty Array\;\\

\For{Every words in RWFrequency}
{
    Add 'word' as index in WeightRWF\;\\
    $WeightRWF[word] = \frac{RWFrequency[word}{lenghtRWF}*100$\;
}

Initial lenghtWRWF $=$ Length of WeightRWF\;

\For{Every words in SWFrequency}
{
    \eIf{word is in WeightRWF}
    {
        $AAScore = AAScore + \frac{WeightRWF[word]}{lenghtWRWF}$\;
    }{
    Do nothing and Continue\;
    }
}

\For{Every words in SWFrequency}
{
  \eIf{word is in RWFrequency}
  {
   $AAScore = AAScore + \frac{SWFrequency[word]}{RWFrequency[word]}*100 + \frac{SWFrequency[word]}{lengthSWF} $\;\\
   $RWFrequency[word] = 0$\;
   }{
   Do nothing and Continue\;
   }
}
\For{Every words in RWFrequency}
{
  \eIf{RWFrequency[word] is not 0}
  {
   $AAScore = AAScore - \frac{RWFrequency[word]}{lengthSWF}*100$\;\\
   $RWFrequency[word] = 0$\;
   }{
   Do nothing and Continue\;
   }
}
\caption{Comparison Algorithm}
\end{algorithm}

\subsection{Final Scoring of Student's Answer}
After comparing and performing linguistic analysis, it will give scores. The marks will be distributed according to the  Table I which can be modified by the admin of the system.

\begin{table}[h]
\centering
    \caption{Score Distribution Depending on Different Steps}
    \label{ScoreDistri}
    \begin{tabular}{ | m{3.5cm}| m{1cm} | } 
        \hline
        Section & Scores \\ 
        \hline
        Frequency of Word & 70\% \\ 
        \hline
        Linguistic Mistakes & 30\% \\ 
        \hline
    \end{tabular}
\end{table}

Using Table I and following formula the final score is given to tha student's answer script.

$FinalScore = TotalMark*0.7*(AAScore/100) + TotalMark*0.3*(LAScore/100)$

\subsection{Simple Illustration with sample Data}
At first, the question will be extracted and searched in the Wikipedia, the process is to make a clear question. For example: What do you know about University of Dhaka?

Now the system will identify the main and unique keywords from the question. For example: "University, Dhaka". Then the data will be searched in the Wikipedia. The example is given below in figure 3: 
\begin{figure}[htb!]
	\centering
	\includegraphics[width=8.5cm,height=5cm]{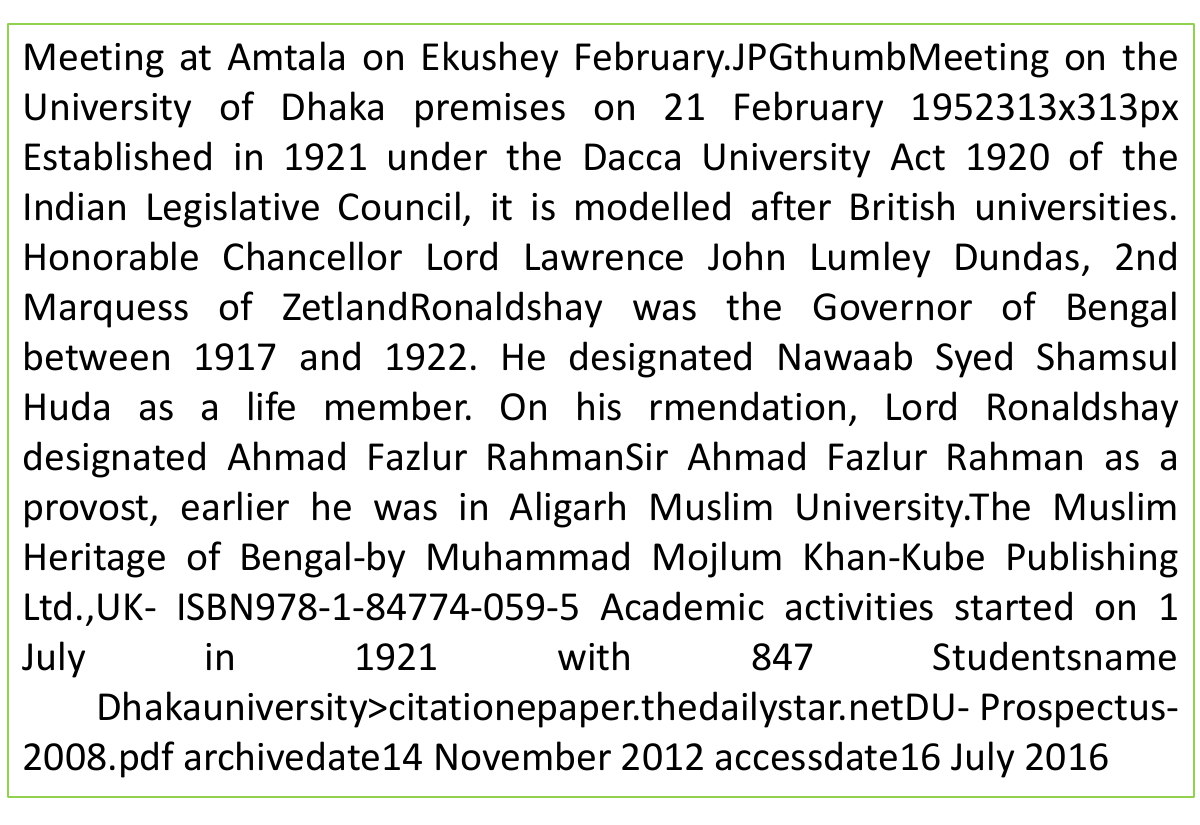}
	\caption{Parsed data about University of Dhaka from Wikipedia}
	\label{fig:fig3}
\end{figure}

Then from the data of the Wikipedia after preprocessing and running frequency calculator algorithm on that data we will get the frequency table and It will be stored as a Model Answer given in figure 4.

\begin{figure}[htb!]
	\centering
	\includegraphics[width=8.5cm,height=4cm]{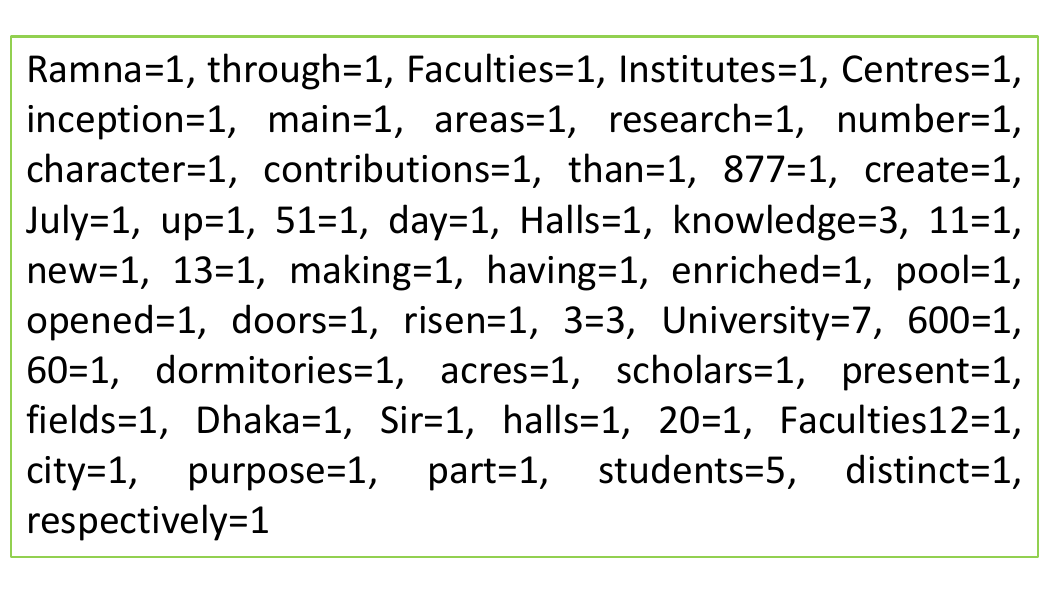}
	\caption{Model Frequency of Parsed Data}
	\label{fig:fig4}
\end{figure}

After that the system will take students answer as a text and consider the following figure 5 as sample.

\begin{figure}[htb!]
	\centering
	\includegraphics[width=8.5cm,height=5cm]{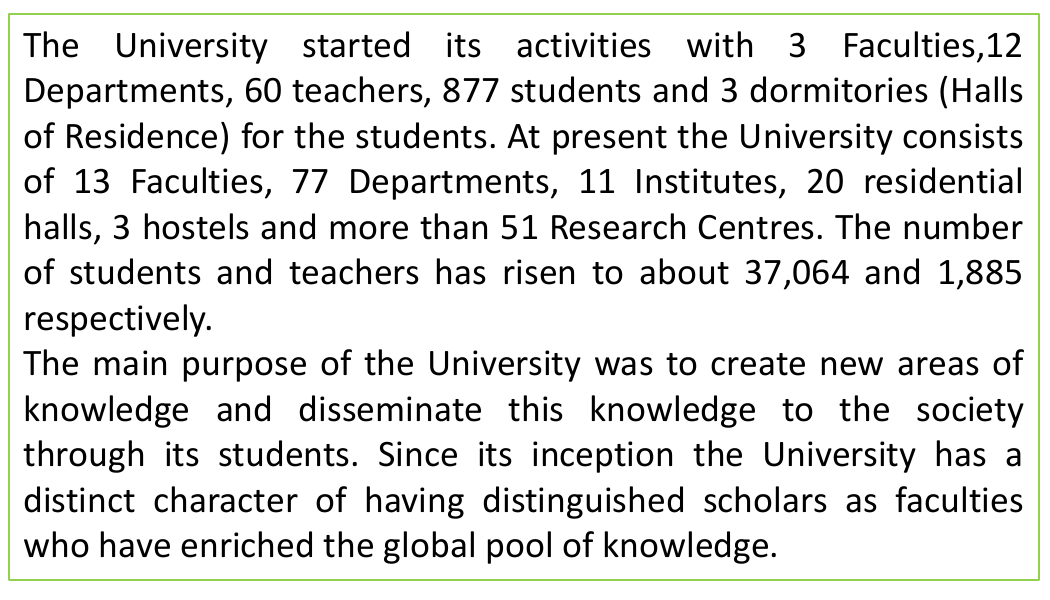}
	\caption{Sample Answer of Student}
	\label{fig:fig5}
\end{figure}

After preprocessing and running frequency calculator algorithm on answer we will get the frequency table and It will be stored as figure 6.

\begin{figure}[htb!]
	\centering
	\includegraphics[width=8.5cm,height=4cm]{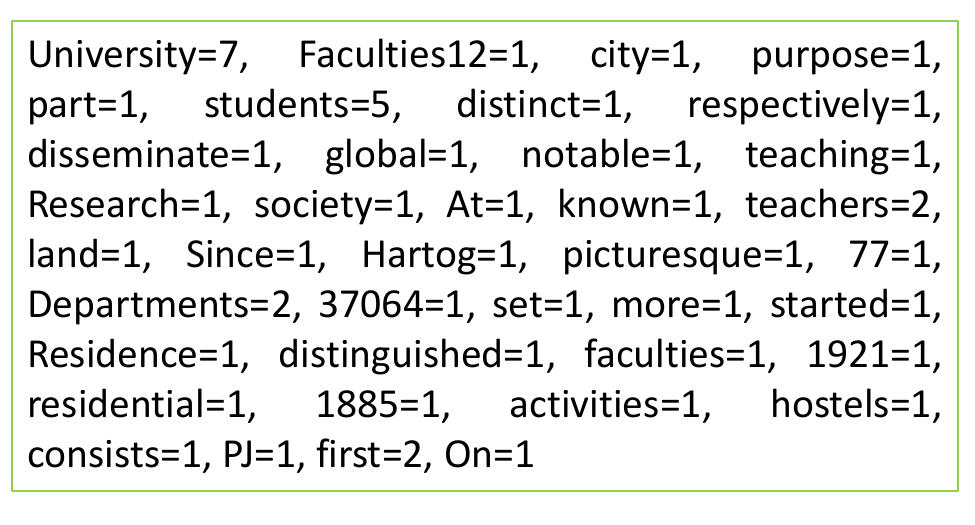}
	\caption{Frequency of Answer of Student}
	\label{fig:fig6}
\end{figure}

The system will compare these frequency tables and give score to the answer applying the comparison algorithm. Finally the score to the student's answer will be evaluated according to the Table I.

\section{Result Analysis}\label{sec:res}
The objective of this research is to automatic evaluation of descriptive answer script and give it a score. To do this we took students answers as text and ran our methodology to test. We took answers from 100 students and tested with our method. We also tested those answer scripts manually 10 teachers of our institute. Then we have calculated Precision, Recall, and F-score for our proposed method of those scores which is given in Table \ref{tab:2}. 

\begin{table}[h]
\centering
    \caption{Precision, Recall, and F-score for our proposed method}
    \begin{tabular}{ | m{1.5cm}| m{3.5cm} | } 
        \hline
         & Score of our proposed system \\ 
        \hline
        Precision & 0.91 \\ 
        \hline
        Recall & 0.81 \\ 
        \hline
        F-score & 0.87 \\ 
        \hline
    \end{tabular}
    \label{tab:2}
\end{table}

From Table \ref{tab:2} we can say that our proposed method is giving a acceptable precision, recall and F-score.

\section{Conclusion \& Future Work}\label{sec:con}
The ease of automatic script evaluation is quite a challenge in modern days. The number of student and job seeker increasing and day-by-day teachers and organizations are facing bigger trouble while trying to evaluate the answer paper. This scenario gives rise to new solutions with the help of the automatic answer script evaluation.  Our system will evaluate the subjective answers. Our system evaluates the student’s answer based on the keywords. Judging upon the model answer and the student’s answer marks will be allocated to the student. Taking an exam with too many students is costly, and evaluating their answer script is time-consuming. The proposed system provides a hassle-free evaluation system in where time and cost are minimized and it decreases human grader errors also. 

\subsection{Future Work}
Since our system is not capable of providing a better efficiency we are proposing some ideas for our future works.
\begin{itemize}
    \item Evaluating answer with a diagram, table, and mathematical expression. After implementing those functions this system will be useful for every evaluation. 
    \item Focusing on the keywords that directly indicate to the question type. For an example: “what”, “when”, “who”, ‘describe”, “Define”.  We know that these keywords indicate to answer related to time, place, person etc. If the system can catch it automatically then processing speed will increase.
    \item It can be applied in the remote learning system. \item Students can give exam from a remote area. 
    \item The current system only evaluates the answers written in English. Further, it can be extended to evaluate answers written in other languages also.
    \item Implementation of a close domain will make it useful for the sensitive exam. 

\end{itemize}
\bibliographystyle{plain}
\bibliography{bib}

\end{document}